\documentclass[twocolumn]{article}
\usepackage{graphicx} % Required for inserting images
\usepackage{multirow}
\usepackage{tabularray}

\title{\vspace{-2.5cm}\textbf{Thinking agents for zero-shot generalization to qualitatively novel tasks}\vspace{.5cm} }
\author{The Obelisk Team\footnote{Correspondance: Thomas Miconi, {\scriptsize \texttt{thomas.miconi@gmail.com}}} \\ Astera Institute\\Emeryville, USA
\vspace{.5cm}
}
\date{} %  Suppress the date

\usepackage{natbib}

\renewcommand{\cite}{\citep}   % Very useful

\begin{document}

\maketitle

\begin{abstract}

Intelligent organisms can solve truly novel problems which they have never encountered before, either in their lifetime or their evolution. An important component of this capacity is the ability to ``think'', that is, to mentally manipulate objects, concepts and behaviors in order to plan and evaluate possible solutions to novel problems, even without environment interaction. To generate problems that are truly qualitatively novel, while still solvable zero-shot (by mental simulation), we use the combinatorial nature of environments: we train the agent while withholding a specific combination of the environment's elements. The novel test task, based on this combination, is thus guaranteed to be truly novel, while still mentally simulable since the agent has been exposed to each individual element (and their pairwise interactions) during training. We propose a method to train agents endowed with world models to make use their mental simulation abilities, by selecting tasks based on the difference between the agent's pre-thinking and post-thinking performance. When tested on the novel, withheld problem, the resulting agent successfully simulated alternative scenarios and used the resulting information to guide its behavior in the actual environment, solving the novel task in a single real-environment trial (zero-shot).

\end{abstract}

\section{Introduction}

An important aspect of intelligence is the ability to handle  novel problems. While simpler organisms are restricted to problems similar to these they have been exposed to during training, and fare badly when faced with novel challenges, intelligent agents can flexibly produce new behaviors to tackle an unexpected situation.  

An major component of this capacity is the ability to \emph{think} before acting. By `thinking'\footnote{Inspired by Berridge's example \cite{berridge1998role}, we specifically use the term `thinking' (with quotes) to  denote the deliberate internal generation, evaluation and selection of actions or sequences of actions. This operationalization seeks to avoid confusion from the many meanings associated with the highly overloaded word \emph{thinking}.}, that is, by internally manipulating concepts and behaviors and evaluating likely outcomes, agents can
tackle novel problems never encountered before, by recombining existing knowledge into new solutions. This ability is perhaps the hallmark of what we think of as truly ``intelligent'' behavior: it is highly prevalent in humans, but is is debated whether it even exists in non-human animals \cite{suddendorf2003mental}, including mammals such as rodents \cite{gillespie2021hippocampal} or even great apes \cite{suddendorf2009great,osvath2010great}.

Much work in machine learning has focused on training agents with increasingly complex innate behaviors. Some work has also focused on making agents more flexible through on-task learning (in-context / few-shot / meta-learning \cite{schmidhuber1987evolutionary,wang2016learning,chollet2019measure}), and/or through the following of arbitrary language instructions \cite{team2021open,team2023human}. Of course, a rigid (but effective) form of `thinking' in the sense of executing fixed, human-defined planning algorithms has long been a staple of machine learning and control research (e.g. model-based control \cite{moerland2023model}). Recently, there have been more proposals for actual `thinking', i.e. the flexible internal generation, manipulation and evaluation of trajectories and outcomes, with varying degrees of success  \cite{guez2019investigation,pascanu2017learning,chung2023thinker,bansal2022end}. Most research in this area tends to test thinking ability with \emph{quantitative} generalization (training on problems up to a certain size, then testing on larger size), within the same general type of problem (mazes, Sokoban, etc.) But the appeal of `thinking' is precisely that it opens the door to true \emph{qualitative} generalization, including to radically different problems (see Related Work for a more detailed discussion of relevant research).

In this work we want to train an agent that makes use of its `thinking' ability to solve a novel problem, never seen before in its training, and without making use of on-task / few-shot learning (i.e. zero-shot)\footnote{Importantly, we are \emph{not} suggesting that `thinking' is superior to, or mutually exclusive with, on-task / few-shot learning. In practice, to address novel tasks, intelligent organisms usually employ a combination of `thinking' and on-line learning (trial and error), and of course constantly use their built-in innate behaviors. However, in this work we want to specifically focus on the `thinking' ability alone, to gain better understanding of its potential and challenges. See Section \ref{sec:behaviors} for more discussion.}.
Furthermore, rather than assessing \emph{quantitative} generalization (on larger versions of the same overall problem), we would like to assess actual \emph{qualitative} generalization: we want a guarantee that the new, test task, on which we evaluate the agent's `thinking' abilities, has not been seen before, even in general principle, while still being solvable solely by `thinking', without additional learning. How can we ensure this?

We notice that most simulated environments involve a number of different interacting \emph{elements}, which may or may not be present in a given episode. For example, in a gridworld-like environment such as Crafter \cite{hafner2022benchmarking}, there are blocks of various types with different properties, tools, trees, cows, zombies, etc. If we withhold a certain \emph{combination} of elements from the training set, while still allowing each element individually, then any new task that critically relies on this specific withheld combination will be genuinely novel: it cannot possibly have occurred during training. Yet, because each element or pair of element (or any partial combination, not amounting to the full withheld combination) was included in the training set, the agent still knows how these elements behave and interact, and can thus generate plans for this new task, without additional on-task learning.

While withholding a combination of elements ensures that tasks involving this combination will be novel, simple novel tasks may still be solvable without `thinking'. Fortunately, whether a task relies on thinking can easily be tested by measuring the difference in performance with and without thinking. See Appendix \ref{sec:reliesonthinking}.

For this approach to succeed, we need to make sure that the agent learns to make appropriate use of its `thinking' ability during training, instead of simply trying to solve all tasks by sheer innate reflexes embedded in the weights. To this end, we ``evolve'' the training tasks to encourage useful `thinking' by the agent: we periodically generate new tasks, then select them for inclusion in the training set based on the difference between the agent's pre-`thinking' and post-`thinking' performance.

\section{Related Work}

As mentioned in Section \ref{sec:behaviors}, behaviors originate from four different broad sources: innate, learned (during lifetimes, on-task, by trial and error), instructed (by another existing agent, through communication or imitation), and planned/forethought (roughly, `thinking' as we use it here). In practice, of course, most actual behaviors involve combinations and interactions of these. Notably, most of machine learning concentrates on the first type of behavior: training innate responses to a given situation (most explicitly in the Markov Decision Process formulation of reinforcement learning).

Generalization to novel situations outside the training set is of course a basic objective of machine learning. One question is how to define what counts as ``novel'', while still being a reasonable challenge. To expand robustness to novelty, one possibility is to train the model to quickly \emph{learn} the relevant aspects of each new task, generally known as ``learning-to-learn'' or ``meta-learning'' \cite{schmidhuber1987evolutionary,wang2016learning}. A prominent example is the ARC challenge, which explicitly defines intelligence the ability to solve novel problems by few-shot learning, and introduces each new task in the form of a small number of training solved examples \cite{chollet2019measure}\footnote{Another observation about the ARC domain is that it relies to a considerable extent on innate knowledge of physics and three-dimensional space, besides generalistic reasoning.}. An alternative possibility is to instruct the new task with language, training the agent to be able to handle arbitrary language instructions - in which case the relevant information about each new task is automatically transmitted through the instruction itself. Outside of dedicated language models, this line of research (going back all the way to SHRDLU \cite{winograd1971procedures}) is today demonstrated in the growing body of work around the XLand project \cite{team2021open,team2023human,nikulin2024xland}.

Planning in general is a well-established field of research, whether in control theory or specifically in reinforcement learning \cite{moerland2023model}. However, this often takes the form of having the agent execute a pre-defined planning algorithm in some model of the environment \cite{todorov2005generalized,silver2016mastering,chua2018deep,hafner2019dream}. 

There are also examples of agents using a world model to imagine possible futures and exploit the resulting information (i.e. `thinking' as we define it here). These works differ in how much pre-defined structure is imposed upon the `thinking' process. At one extreme, ``model-free planning'' simply lets a multi-layer recurrent network run multiple steps for each actual time step in the environment \cite{guez2019investigation}. Imagination-Based Planning \cite{pascanu2017learning} maintains a tree of imagined future states and uses a separate trained module to choose which node in this tree should be used for future imagination. At the other extreme, Thinker \cite{chung2023thinker} condenses information from imagined steps into a set of summary statistics to guide both further imagination and action. 
A special form of ``learning to think'' is when the agent is trained explicitly to produce a certain\emph{algorithm} (embedded in the learned weights of an appropriate architecture), making it automatically able to solve any instance of a given problem, of any size \cite{kaiser2015neural,ibarz2022generalist,bansal2022end}. 

We note that these planning/`thinking' models tend to be tested on \emph{quantitative} generalization: test problems larger than the training set, but still in the same domain (Sokoban, mazes, lander control, etc.). In this work we propose one way of defining \emph{qualitatively} novel tasks, that are still theoretically solvable zero-shot by `thinking' agents.

Here we adopt an intermediate position of organizing `thinking' as agent-controlled rollouts in the world model. A drawback of this approach is its potentially large computational cost, as already pointed out by \citet{schmidhuber1990making}. However, it also has the advantages of simplicity and interpretability, and allows us to use the actual environment simulator as a drop-in replacement for the thinking process. Discovering more efficient forms of `thinking' is an important direction for future work.

Note that `thinking' here is different from the  use of world models to provide a next-step prediction as additional input to a control policy \cite{schmidhuber1990making,ha2018recurrent}, or as surrogate environments to train a policy in \cite{sutton1991dyna,hafner2019dream}.

Recently, large language models have shown remarkable abilities in all four types of behaviors: immediate responses, in-context learning, instruction following, and even spontaneous reasoning. While these models are generally restricted to the domain of language, there are ongoing efforts to connect them with the physical environment \cite{brohan2023rt,driess2023palm}.

\section{Methods}

\subsection{Environment}

The environment is a simple grid world, which can be thought of as a simplified version of Crafter \cite{hafner2022benchmarking}. At each discrete location, the world may contain the agent, a block, or a non-playing character (``mob''). Mobs include zombies, which noisily follow the agent and cause negative reward when touching the agent; cows, which move randomly and that the agent can kill to obtain positive reward; and ``angels'', which move randomly and kill zombies on contact but do not otherwise interact with the agent. Blocks include unmovable ``stone'' blocks, ``pickable'' blocks that the agent can pick and drop,  ``killer'' blocks which the agent can pick and drop on killable mobs and blocks (killing them and consuming the killer block), ``diggable'' blocks that cannot be picked, but can be destroyed by dropping a killer blocks on them, and ``fruit trees'', which when touched by the agent provide a one-time reward and become ``empty trees'' for a fixed period of time before sprouting fruit again. Cows, zombies and angels can all be killed by dropping a killer block on them. All blocks (other than non-diggable stone) can also be destroyed by dropping a killer block on them. Importantly, killer blocks are single-use and disappear when successfully killing a mob or block. The agent gets positive reward for touching a fruit-bearing tree, killing a cow, or when a zombie dies (either being killed by the agent or by an angel). The agent gets negative reward when touched by a zombie.

\begin{figure}[t]
    \centering
    \includegraphics[width=0.4\linewidth]{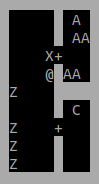}
    \hspace{.5cm}
    \includegraphics[width=0.4\linewidth]{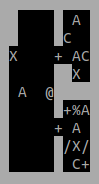}
    \caption{Environment tasks (levels). The agent is represented as `@'. Left: The test task, based on a combination of Zombies (Z), zombie-killing angels (A) and killing/digging blocks (X). Right: Random training task. Notice the presence of Angels (A) and digging blocks (X), but the absence of Zombies (Z). No training task can contain all three of Z, A and X.}
    \label{fig:levels}
\end{figure}

Our environment is implemented in fully vectorized PyTorch. This allows us to run a large number of environments in parallel with very little increase in compute time (experiments in this paper use a batch size of 5000). Thus we can run the agent on multiple tasks, with multiple runs per tasks, all in parallel in a single batch. This ability is crucial to efficient evaluation not just of the agent's behavior, but also of task characteristics, as explained below.

\subsection{Agent structure}

The agent itself is a simple recurrent network, endowed with a separate world model that it uses for simulating trajectories in the environment. For the present paper, for practical convenience, we chose to keep the world model as separate network, and also to train it separately from the agent (but on the same distribution of tasks, i.e. respecting the withheld combination - see below). The world model takes as input a current observation and an action, and outputs a stochastically sampled prediction for the next observation. Thus, it can be used for simulating trajectories in the environment, without any actual environment interaction taking place. See Section \ref{sec:agent} for details.

\subsection{Training}

Each episode consists of an initial observation at time $t=0$ from the environment, a fixed period of `thinking' (with no environment interaction), and then a single trial of 30 timesteps in the environment. For the present introductory paper, we chose to provide strong structure to the `thinking' process, which consists of exactly 3 simulated trials. During each simulated (`thinking') trial, the agent starts with the initial observation from the true environment and then proceeds to simulate a trajectory, choosing its own actions and using its world model to predict a next (reconstructed) observation, for three full trials; at the start of each such internally-simulated trial, the current observation is reset to the true environment's initial observation. However, the hidden state of the agent's recurrent network is \emph{not} reset between trials (`thinking' or ``real''); thus, at any time, the agent may maintain and exploit memories off all previous timesteps and trials in the current episode. A special input indicates to the agent whether it is `thinking' (first three simulated trials) or acting in the actual environment (last, ``real'' trial). Again, we stress that no actual interaction with the real environment occurs during `thinking': only after the `thinking' process is complete does the agent run a single trial of 30 steps in the actual environment. The agent's final score for the episode is the total net reward accumulated during the final, ``real'' trial (reward from the `thinking' trials is not used for agent score, though as explained below it is involved in selecting useful tasks). 

Another important simplification for this paper is that, while training the agent, the `thinking' trials do not occur in the agent's own world model but in the actual environment simulator, reset for each trial. In other words, during training, the agent ``thinks'' in the simulator rather than in its world model. Only for the test trials, on the test task, does the agent use the actual (separately trained) world model for the three `thinking' trials:  the world model is used as a drop-in replacement for the environment simulator. This choice allows us to train the agent separately from the world model, then join them together at test time. We stress that this choice was dictated by engineering considerations (see Appendix).

A task is entirely defined by the initial configuration of the world: the mechanics of the environment are fixed. To define a task, one simply specifies the mob or block occupying each position at time zero.
Note that,  from a given initial observation, and even without considering agent actions, the environment may evolve into entirely different futures due to stochasticity (e.g. in the mob's behaviors).
The agent is initially trained for 1000 episodes on a set of 10 hand-defined starting tasks, which are expected to equip the agent with certain basic skills (picking and dropping blocks, killing zombies or cows). After this initial period, we periodically generate new tasks, have the agent run a full episode on them (including  three `thinking' trials and one actual trial), and select the 10\% of tasks that get the highest ``task score'', defined as the mean difference in agent score between the last and first trial (note that each task is replicated multiple times in the batch dimension, which smoothes out the stochasticity from mob and agent actions). Since we train in the environment simulator itself, first trial reward is a faithful estimation of the ``pre-thinking'' agent performance, while  last trial reward represents ``post-thinking'' performance. These selected tasks are then included into the training set, replacing the lowest-scoring tasks from the current set. See Section \ref{sec:agent} for details.

During the entire training process (for both world model and agent) we strictly prevent the occurrence of one specific combination of elements in any task. This withheld combination of elements, in turn, is crucial to the test task. 

\subsection{Test task}

The test task consists of an open space with two rooms. Each room is enclosed by undiggable stone, and has a ``door'', composed of a single diggable block. In the open space, there is one killer block (which can also be used for digging), and also a number of zombies. One of the two rooms contains a single cow, as a distractor. The other contains a number of angels (which can kill zombies on contact). See Figure \ref{fig:levels}, Left for a visualization.

The agent starts at the center of the world, in the open space. Thus, if it fails to act correctly, it will likely be harassed by the zombies and incur many negative rewards. While it may use the (single-use) killer block to kill one zombie, it will then be unarmed and unable to fend off the remaining zombies. The correct solution is to use the killer block, not to kill a zombie, but to dig out the door to the room containing the angels. This will allow the angels to escape the room, enter the open space, and wipe out the zombies, removing their threat and accruing positive reward to the agent. In particular, the agent must not get distracted by the cow in the other room, which would bring no benefit to the agent. Note that there is only one killer block (which, again, is single use and disappears upon successful use), so the agent can only dig out one of the two doors, or kill one zombie. 

By itself, this task is very simple, and a typical reinforcement learning agent could learn it quickly from training. However, we specifically want to prevent this task, in any form, from appearing in the training set. Our goal is to ensure that this task remains truly novel, even in broad principle, when it is encountered for the first time during the test episode.

Note that the optimal solution to the task relies on a specific combination of elements: zombies, angels, diggable blocks (doors), and killer blocks (used for digging). If any task does not include one of these elements, a solution for this other task cannot be applied as-is to the test task. Therefore, during training, we strictly prevent any task from containing all of these elements together\footnote{In fact, for agent training, we actually use a stricter standard and withhold the combination of zombies, angels and killer blocks.}. Importantly, each individual element, or even any partial combination of them, is allowed, which allows the agent to learn the behavior and interactions of the elements: only the full set is prevented from appearing together in any training task (both for the agent and the world model). Thus, at test time, when the agent encounters the test task for the first time, it cannot simply apply an already-learned innate behavior. Rather, it must make use of its `thinking' period to test possible courses of actions in the world model, and guide behavior during the single ``real'' trial.

\section{Results}

\subsection{Learning to `think'}

The main objective of this work is to build an agent capable of using its `thinking' ability (internally generating and evaluating possible trajectories) to solve a qualitatively novel problem in a single real trial. Our main measure of success is the difference between pre-thinking and post-thinking result on a test task, based on a withheld combination of elements which ensures qualitative novelty. To assess pre-thinking, or during-thinking behavior, we make use of the fact that `thinking' in this experiment takes the form of internally simulated rollouts, which can be reproduced in a copy of the actual environment.

\begin{figure}[t]
    \centering
    \includegraphics[width=.95\linewidth]{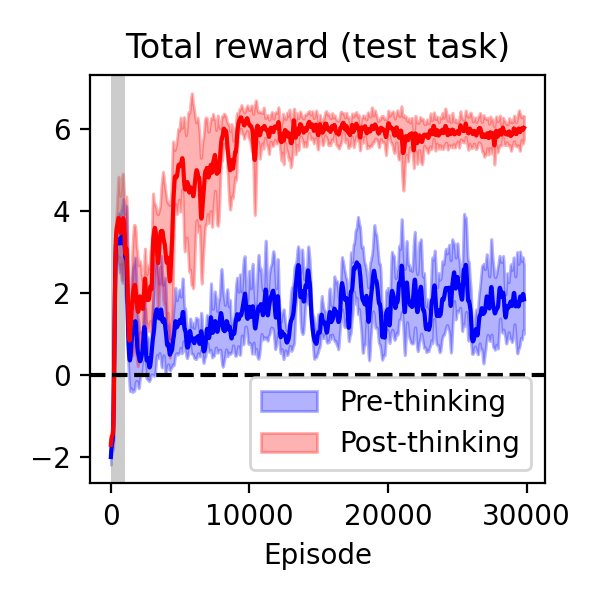}
    \caption{Total reward on the novel test task, for trial 1 (first `thinking' trial), and trial 4 (single real-environment trial). Gray shaded area represents the pre-training period on a fixed set of hand-defined tasks; training after this includes randomly generated, selected tasks. }
    \label{fig:results}
\end{figure}

Figure \ref{fig:results} reports the total reward accumulated over a trial on the test task, both for the first (pre-thinking) trial and the last (post-thinking) trial, over the course of training (ground-truth rewards are obtained for the first, `thinking' trial by simulating the internally generated actions in a copy of the actual environment). After a joint increase during the pre-training period, pre-thinking performance decreases, while post-thinking performance increases, reflecting the successful emergence and exploitation of `thinking' ability. Note that the plateau performance is quite below the optimum, which is 8.0 for this particular task (four zombies being killed by angels, for a reward of 2.0 each). 

\begin{figure}[t]
    \centering
    \includegraphics[width=.95\linewidth]{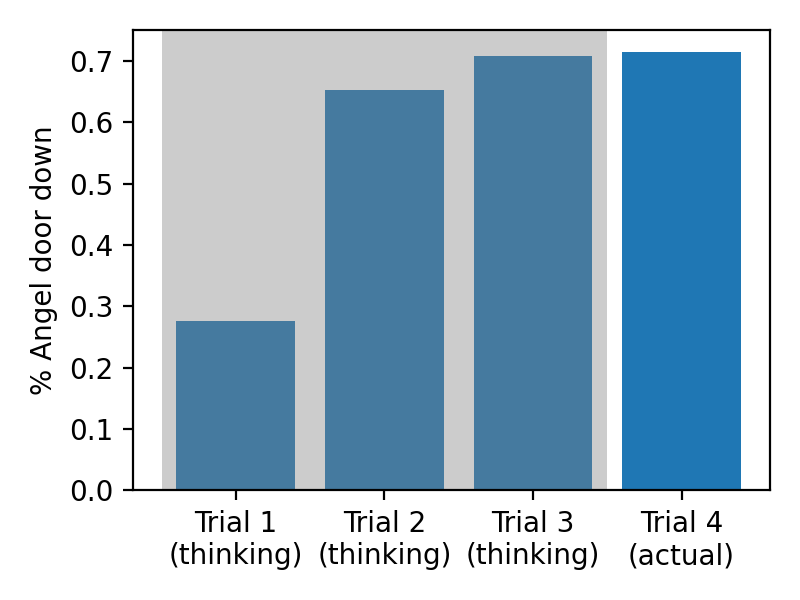}
    \caption{Test task: proportion of test episodes in which the angel door is opened, in a batch of 5000 test episodes with the fully trained agent. For `thinking' episodes (gray shaded area), this is assessed by running the internally generated actions in a copy of the actual environment (the agent's perceived observations during `thinking' are always internally generated from its own world model, with no interaction with the actual environment).}
    \label{fig:angeldoordown}
\end{figure}

The known optimal strategy on this test task is to break open the door of the room that encloses the zombie-fighting ``angels''. We therefore assess the thinking process of the agent by computing the probability that this door gets broken, at the end of each `thinking' trial (again by replicating internally generated action in a copy of the actual environment), and in the single actual trial. We run the fully-trained agent on a batch of 5000 environments on a single test episode, and plot the proportion of environments (out of 5000) with the angel door broken down at the end of each trial (note that while all test episodes share the same overall configuration, the actual position of the angel door is randomized by random flip for each environment). As shown in Figure \ref{fig:angeldoordown}, in its initial `thinking' trial the agent rarely chooses to break down the angel door. However, over successive trials, the benefits of freeing the angels are high enough that the agent quickly settles on this course of action.

\subsection{Additional Experiments}

\subsubsection{Causal influence of `thinking' on final behavior}

Because `thinking' in our model occurs as explicit mental rollouts, we can manipulate what the agent perceives during the `thinking' period. This allows us to determine whether the agent truly and flexibly makes use of its imagined behaviors, rather than being mostly driven by ``lucky'', innate biases. For these experiment, we again perform the `thinking' in the simulator, rather than the agent's world model, allowing us full control of the agent's perception during `thinking'.

We run the exact same test task as above, on the exact same (frozen) trained agent, but manipulate a crucial parameter during the `thinking' period - namely, the reward received by the agent when an angel kills a zombie.  We vary this reward, and observe the impact on the agent's behavior during the last trial. As reported in Table \ref{table:akzreward}, we find that setting this reward to zero reduces the probability that the agent will choose to open the angels' door in the last, ``true'' trial. Furthermore, making this reward negative (i.e. the agent gets punished with negative reward if an angel kills a zombie) reduces this probability even more, though not all the way to zero.  This demonstrates that the agent does flexibly make use of information gathered during its `thinking' period to control its behavior during the final, ``real'' trial.

\begin{table}
\begin{center}
\begin{tabular}{|c|c|}
\hline
\textbf{AKZ Reward} & \textbf{P(Angel Door Opened)}  \\
\hline
 2.0 (default)   & 0.86  \\
 \hline  0.0   &  0.61 \\
 \hline -4.0 &  0.25  \\
\hline
\end{tabular}
    \caption{Testing the causal influence of `thinking' on behavior. Manipulating the perceived reward when an Angel kills a Zombie (``AKZ reward'') during the thinking period changes the probability that the agent will choose to open the angels' door during the last, ``true'' trial. This demonstrates that the agent does flexibly make use of information gathered during its `thinking' period. Note that the top-right number is analogous to the rightmost number in Figure \ref{fig:angeldoordown_inenv}.}
\label{table:akzreward}
\end{center}
\end{table}

\subsubsection{Ablations}

Further experiments allow us to identify which components of the system most affect performance. In the Appendix, we show that an important bottleneck of the present system is the world model, more specifically the next-latent predictor: when using the actual environment simulator for `thinking' (rather than the agent's own world model), the proportion of correct door choice jumps to $85\%$. See Section \ref{sec:ablations} for details.

\section*{Acknowledgements}

The Obelisk team at the Astera Institute is Thomas Miconi, Kevin \mbox{McKee}, Yicong Zheng, Sam Nolen (researchers), Gary Miguel, Andrew Grebenisan, Hong Xu, Eric Alt, Mick van Gelderen, Matt Behrens (engineering) and Jed McCaleb (manager).

\bibliography{biblio}
\bibliographystyle{abbrvnat} % name, year
%\bibliographystyle{unsrt} % numbers

%\appendix
%\onecolumn

\renewcommand\thefigure{A\arabic{figure}}    
\setcounter{figure}{0}    

\renewcommand\thesection{A\arabic{section}}    
\setcounter{section}{0}

\section*{Appendix}

\section{Sources of behaviors}
\label{sec:behaviors}

Where do behaviors come from? When facing any given situation, an agent can draw its behavioral responses from at least four different origins: \emph{innate} (imprinted in genotype and structural phenotype by evolution), \emph{learned} (from lifetime experience and stored in synaptic connections), \emph{instructed} (through imitation of, or  explicit command from, another agent), or \emph{planned/forethought} (internally conceived and deliberately selected after estimation and evaluation of likely outcomes) (see Table \ref{table:behaviors} for a summary). 

In practice, of course, most non-reflex behaviors result from interactions between these four sources: planned or instructed behaviors will eventually be repeated and learned if found useful, innate behaviors and lifetime learning define a repertoire from which deliberate planning can pick, innate tendencies induce biases in learning outcomes, learning may eventually guide the course of evolution through the Baldwin effect \cite{hinton1987learning}, etc. In addition, obviously all these abilities arose from evolution in the first place. Nevertheless, identifying these distinct sources of behavior may facilitate a better understanding of intelligent behavior.

Of these four sources of behavior, the fourth (deliberate forethought and planning, hereafter referred to as `thinking' is perhaps the most strongly associated with what we think of as ``intelligent'' behavior. Although commonly observed in humans, it is still debated whether it occurs in related mammals such as rodents, or even monkeys. 

Importantly, `thinking' before acting allows an agent to tackle novel problems never encountered before, by recombining existing knowledge into new solutions. Relying on innate behaviors alone only works on problems highly similar to those seen during training (evolution). On-line learning requires random trial and error, which is slow, often impractical, and potentially dangerous in the real world. By devising potential trajectories and evaluating their likely outcomes internally, we can come up with novel solutions to novel problems.

\begin{table*}
\begin{center}
\begin{tabular}{|c|c|}
\hline
\textbf{Behavior type} & \textbf{Source}  \\
\hline
 Innate   & Evolution  \\
 \hline  Learned   &  Lifetime learning  \\
 \hline Instructed &  Imitation and direction\\
 \hline \multirow{2}{*}{Planned}  & Internal generation, evaluation and selection of \\ & actions or sequences of actions (`thinking')\\
\hline
\end{tabular}
    \caption{Four difference sources of behaviors. In practice, actual behaviors result from interactions and combinations of these four components.}
\label{table:behaviors}
\end{center}
\end{table*}

\section{Novelty and Thinking}
\label{sec:reliesonthinking}

Novel tasks (including tasks relying on withheld combinations of elements) do not necessarily require `thinking' if they are very simple. For example, if the novel task simply requires addressing each element in isolation, with no significant interaction between the elements, then the innate behaviors acquired during training (during which the agent learned about each individual element) will be sufficient to address this novel task. Thus, it is important that the novel task should explicitly rely on the interaction of all elements in the withheld combination for success. 

It may not always be straightforward to determine \emph{a priori} whether this condition holds for a given task. Fortunately, using the methods in this paper, we can directly assess difference in performance between pre-`thinking' and post-`thinking' trials. Consistent increases indicate that the task is complex enough to benefit from `thinking'.

\section{Agent structure and training}

\label{sec:agent}

The agent is a Long Short-Term Memory recurrent network \cite{hochreiter1997long}, trained by a simple Advantage Actor-Critic algorithm \cite{mnih2016asynchronous} over the entire episode of 120 steps (4 trials of 30 time steps each). Only reward from the last trial is taken into account.

Because our environment is written in fully vectorized PyTorch, iteration time is almost independent of batch size, allowing us to use large batches. We use a batch size of 5000, divided into 100 different tasks (each repeated 50 times). This allows us to obtain a good estimate of the expected reward for a given task.

The observation space consists of two $9\times9$ binary grids, one for characters (``mobs'': the agent itself, zombies, angels, cows) and ones for blocks (stone, diggable blocks, pickable blocks, killer/digging blocks, fruiting trees, empty trees), as well as a numeric input for previous-step reward, a one-hot encoding of the previous action, a timer indicating the proportion of total episode time spent, an indicator for whether the current trial is the last (``real'') one (as opposed to a ``thinking'' one), a one-hot encoding (for each block type) about whether the agent is currently carrying a block of that type (the agent can carry at most one block at any time), and an end-of-trial indicator. The total size of the observation vector is 837 (including a few unused inputs), most of which is binary inputs except for timer and previous reward, which are real-valued.

The action space consists of no-operation, moving in any of 4 cardinal directions, and dropping a currently-carried block in either of the 4 cardinal directions, resulting in a single one-hot vector of size 9. The agent's ``orientation'' is irrelevant to the dynamics (unlike Crafter). To pick a block, the agent simply moves towards the square on which this block is located; the block is picked (and the agent moves to this block) if it is pickable and the agent is not currently carrying another block, otherwise no change or movement occurs. The agent can drop a block it is currently carrying if the target square is empty, or if it is occupied by a killable object (diggable wall, tree or mob) and the carried block is a killer/digging block, otherwise no change occurs. If the killing block is dropped onto a killable object, the target square is cleared and the block vanishes (i.e.    killing/digging blocks are single-use).

At any time, the batch of size 5000 is divided into 100 tasks, each replicated 50 times. The agent is initially trained for 1000 episodes on a set of 10 hand-defined starting tasks, which are expected to equip the agent with certain basic skills (picking and dropping blocks, killing mobs, etc.). After this initial period, we periodically generate new tasks and evaluate them for inclusion in the current training set. The ``score'' of a task is the difference between last-trial reward and first-trial reward, mean-centered and divided by one plus the standard deviation of scores over the 50 replicates of the task. During each task generation event, a whole new batch of 100 tasks is randomly generated and evaluated for one episode, then the 10\% of current-set tasks with the lowest scores are replaced with the best-scoring newly generated tasks. Thus the set of tasks on which the agent is trained evolves slowly over time. Since we train in the environment simulator itself, first-trial reward is a faithful estimation of the ``pre-thinking'' agent performance, while  last-trial reward represents ``post-thinking'' performance. This choice of task score is expected to favor tasks that encourage the agent to develop `thinking' ability, that is, explore mentally during `thinking' trials and make use of resulting information in the final trial. 

We stress that the task score is only used for selecting and replacing tasks in the training set. The agent is only trained based on last-trial reward alone.

\section{World model}

A ``world model'' here is defined as a function that takes in a current state and an action, and outputs a probability distribution over possible next observations (or a sample from this distribution).

Training a world model is straightforward for deterministic, fully observable environments. However, it becomes a considerably harder problem in stochastic, partially observable environments:
\begin{itemize}
    \item Stochasticity leads to failure modes such as the ``blurring problem'': when one starting point leads to multiple possible distinct futures, the model may instead output a merged or blurred interpolation between those possible states. 
    
    \item Partial observability requires the memorization and recall of information that momentarily exits the observation field. A common stress test of world models, especially for three-dimensional environments, is to make the agent spin 360 degrees in the model and compare the starting observation with the final one.

\end{itemize}

A common framework for world models is to divide them into a state encoder, which summarizes the current state into a lower-dimensional latent vector (hereafter just ``latent''), and a next-latent predictor, which outputs a probability distribution over possible next-step latents.

In this work, rather than encoding each individual state separately, we chose instead to encode transitions between states. That is, given two successive states $s_t, s_{t+1}$, and the intervening action $a_t$ and reward $r_t$, we train an encoder $E$ to output a latent $l = E(s_t, r_t, a_t, s_{t+1})$, such that $l$ should contain enough information to usefully reconstruct $s_{t+1}$ and $r_t$ when given only $s_t$ and $a$. Note that this component is strictly deterministic for any given $s_t, a_t, r_t, s_{t+1}$ quadruplet. We train this component as a simple binary auto-encoder (\emph{not} variational), that is, an auto-encoder in which the intermediate latent vector $l$ is constrained to  be composed of binary 0 or 1 elements (with gradients transmitted from the decoder $D$ to the encoder $E$ via a straight-through estimator).

Although developed independently, our method of encoding transitions instead of individual observations is exactly similar to the one proposed by \citet{micheli2024efficient}.

The main reasons for encoding transitions rather than individual states are, first, that the changes between successive states are often rather small compared to the description of the whole state; second, given a pair of states and an action, all the deterministic aspects of the state-to-state transition can now be stored in the weights of the decoder itself, leaving only the non-deterministic components to be stored in the latent $l$.

The latent transition encodings $l$ are binary vector of size $12$, so there are $4096$ possible latents in total. This allows us to model the next-latent prediction as a multinomial (categorical) probability distribution over $2^{12}=4096$ possible next latents, similar to language models outputting a probability distribution over all $N$ possible tokens. By sampling from a probability distribution over all separate latents, rather than reconstructing a possible next latent, we hope to alleviate the ``blurring problem'' mentioned above. 

For our simple grid world, observations are binary vectors of size 820. The transition encoder takes two successive observations, as well as an action embedding (one-hot vector over 9 possible actions) and an embedding of reward obtained during this transition (a single floating point value between $-10.0$ and $10.0$), concatenates them, pass them through two residual blocks of hidden size 800, then through a bottleneck of size 12 passed through a sigmoid function, resulting in activations $z$. $z$ is then binarized by a hard threshold at 0.5, resulting in the actual binary latent $l$. The decoder takes this binary latent, together with the initial observation and action embedding, passes them through two residual blocks of hidden size 800, and outputs a reconstructed next-step observation and reward. To make use of the straight-through trick, the binary latent is actually passed to the decoder as \verb|l.stop_gradient() + z - z.stop_gradient()|, as suggested by \citet{hafner2020mastering}. The loss is plain mean-squared error between the predicted and actual next-step observation and between-steps reward.

Again, we reiterate that the transition encoder is strictly deterministic: For any transition $(s_t, a_t, r_t, s_{t+1})$, the encoder $E(s_t, a_t, r_t, s_{t+1})$ outputs exactly one deterministic binary latent $l$, and similarly the decoder $D(s_t, a_t, l)$ outputs one deterministic reconstructed $\hat{s}_{t+1}$ and $\hat{r}_t$. Handling the non-determinism of the environment is the role of the next-latent predictor $P(s_t, a_t)$

The next-latent predictor $P$ is a Long Short-Term Memory recurrent network \cite{hochreiter1997long}, of hidden size 500. At each time step, it takes as input a previous observation, a one-hot embedding of the action (passed through two residual block of hidden size 500), and outputs (after two linear layers) the logits of a probability distribution over all $4096$ possible binary latents. One latent is sampled from this distribution, and decoded through the trained latent decoder discussed in the previous paragraph, resulting in a predicted next-step observation and reward $\hat{s}_{t+1}, \hat{r}_t$. $P$ is trained by cross-entropy loss over the latents, with the target latent being the one actually produced by the encoder $E(s_t, a_t, r_t, s_{t+1})$ given the true transition. $P$ is a recurrent network because some memory of previous time steps is likely to be useful, due to the partial observability of the environment. Unlike the agent's control policy, $P$'s hidden state is actually reset between trials of a given episode: no information needs to be carried across trials for within-trial next-step prediction.

In practice, as explained in the main text, we train the agent, the latent encoder and the next-latent predictor separately. During training, the agent performs its `thinking' in the actual environment simulator. After the agent is trained, we generate new episodes to train the latent encoder and decoder $E,D$. Then, we further generate more episodes to train the next-latent predictor $P$. Note that during this entire training procedure, we scrupulously prevent the appearance of the withheld combination (Zombie, Angel, X-block) during any given episode.

\section{Ablations}

\label{sec:ablations}

Because our method relies on simulated rollouts, we can run the agent while individually ablating  various components of the system. 

\begin{figure}[t]
    \centering
    \includegraphics[width=.95\linewidth]{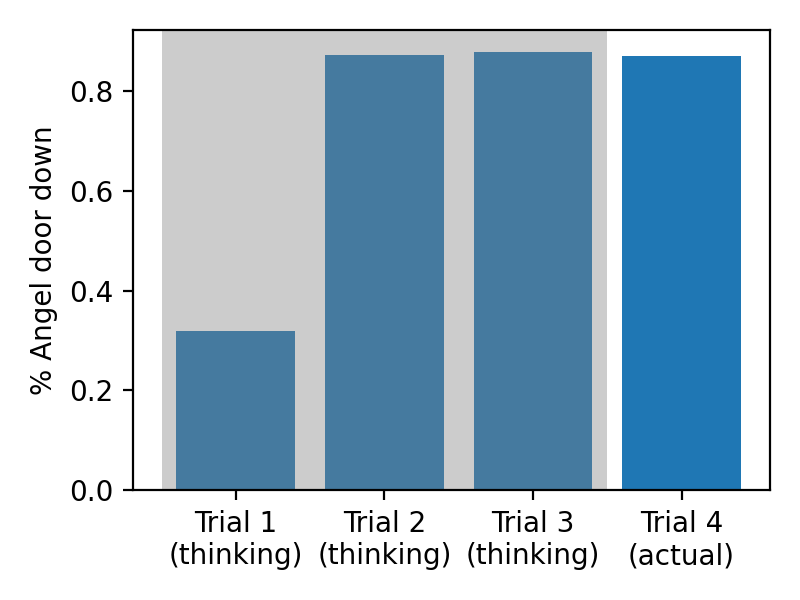}
    \caption{Proportion of test episodes with the angel door down, but replacing the agent's world model with the true environment simulator (conventions are as in Figure \ref{fig:angeldoordown}).}
    \label{fig:angeldoordown_inenv}
\end{figure}

First, we ran the trained agent on the test task, but by using the actual true simulator rather than the world model for `thinking'. Unsurprisingly, when given the true observations and rewards resulting from its imagined actions, the agent performs significantly better. As shown in Figure \ref{fig:angeldoordown_inenv}, the agent now chooses to open the angel door about $85\%$ of the time in the final trial. This suggests that

\begin{figure}[t]
    \centering
    \includegraphics[width=.95\linewidth]{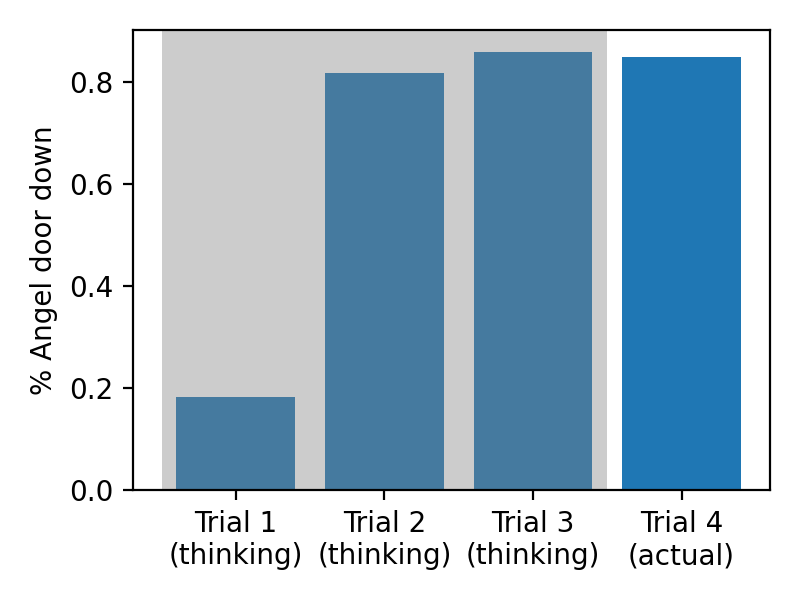}
    \caption{Same as Figure \ref{fig:angeldoordown_inenv}, but passing the observations through the agent's latent encoder. High performance suggests that most of the loss when thinking in the world model (compare Figure \ref{fig:angeldoordown}) comes mainly from the next-latent predictor, rather than information loss in the latent.}
    \label{fig:angeldoordown_inimg_truelatents}
\end{figure}

The world model is composed of a latent encoder, and a next-latent predictor. Can we separate their respective impacts on performance? We again let the agent ``think'' in the simulator, but now instead of providing the true observation from the environment, we pass the current observation and the next-step observation through the latent encoder, obtain a latent, and use it to reconstruct the next-step observation. This is equivalent to `thinking' in the world model, but with optimal latent prediction. Performance is comparable to that obtained by `thinking' entirely in the simulator (figure \ref{fig:angeldoordown_inimg_truelatents}), suggesting that the main difficulty with the world model arises from correct next-latent prediction rather than losses in the latent encoding.

\end{document}